\documentclass[final]{l4dc2026}

\title[CoFineLLM: Conformal Finetuning of LLMs for Language-Instructed Robot Planning]{CoFineLLM: Conformal Finetuning of Large Language Models \\for Language-Instructed Robot Planning
}
\usepackage{times}
\usepackage{balance} 

\usepackage{enumitem}
\usepackage{mathtools}

\usepackage{algorithm}
\usepackage{algorithmic}
\usepackage{multirow}
\usepackage{mysymbol}
\usepackage[title,titletoc]{appendix}
\usepackage{wrapfig}
\usepackage{booktabs}

\coltauthor{\Name{Jun Wang} \Email{junw@wustl.edu}\\
 \Name{Yevgeniy Vorobeychik} \Email{yvorobeychike@wustl.edu}\\
 \Name{Yiannis Kantaros} \Email{ioannisk@wustl.edu}\\
 \addr Washington University in St. Louis, 1 Brookings Drive St. Louis, MO 63130}

\begin{document}

\maketitle

\begin{abstract}%
Large Language Models (LLMs) have recently emerged as planners for language-instructed agents, generating sequences of actions to accomplish natural language tasks. However, their reliability remains a challenge, especially in long-horizon tasks, since they often produce overconfident yet wrong outputs. Conformal Prediction (CP) has been leveraged to address this issue by wrapping LLM outputs into prediction sets that contain the correct action with a user-defined confidence. When the prediction set is a singleton, the planner executes that action; otherwise, it requests help from a user. This has led to LLM-based planners that can ensure plan correctness with a user-defined probability.
However, as LLMs are trained in an uncertainty-agnostic manner, without awareness of prediction sets, they tend to produce unnecessarily large sets, particularly at higher confidence levels, resulting in frequent human interventions limiting autonomous deployment.
To address this, we introduce CoFineLLM (Conformal Finetuning for LLMs), the first CP-aware finetuning framework for LLM-based planners that explicitly reduces prediction-set size and, in turn, the need for user interventions. We evaluate our approach on multiple language-instructed robot planning problems and show consistent improvements over uncertainty-aware and uncertainty-agnostic finetuning baselines in terms of prediction-set size, and help rates. Finally, we demonstrate robustness of our method to out-of-distribution scenarios in hardware experiments.
\end{abstract}


\section{Introduction}

Task planning is a fundamental capability in robotics, enabling autonomous agents to design sequences of actions to accomplish high-level missions. Such missions are typically specified using formal languages \citep{sahin2019multirobot,kantaros2020stylus,ye2024prioritize,kamale2024optimal,vlahakis2025conformal} or optimization-based formulations \citep{chen2017socially,kiran2021deep,zhou2023identify,sachdeva2024uncertainty,pmlr-v283-wang25d}. These approaches, while powerful, require significant user expertise and manual effort to craft correct task specifications.\footnote{Our software implementation of CoFineLLM is available at \href{https://cofinellm.github.io}{https://cofinellm.github.io}.}

In contrast, natural language (NL) offers a more intuitive interface for specifying robot missions. Motivated by the strong generalization abilities of large language models (LLMs), recent work has introduced a new paradigm of high-level planning from NL instructions \citep{ahn2022can,singh2023progprompt,shah2023lm,ding2023task,liu2023llm+,yang2024text2reaction,rana2023sayplan,garg2024large,wang2024llm,li2025llm,zhang2025namo,shi2025pip,ravichandran2025deploying,schwartz2025efficient,strader2025language,xu2025nl2hltl2plan}. Despite impressive empirical results, LLM-based planners often hallucinate and generate overconfident yet wrong actions. 
%
To mitigate this, recent works have leveraged Conformal Prediction (CP) \citep{angelopoulos2023gentle,shafer2008tutorial} to quantify uncertainty in LLM-based planners \citep{ren2023robots,wang2024safe}. The planning problem is cast as a sequence of multiple-choice question–answering (MCQA) tasks, where the LLM is queried to select an action from a finite set. Instead of selecting the action with the highest model's confidence score, these works apply CP to produce a prediction set guaranteed, with user-specified confidence, to contain the correct action. When the set is a singleton, the planner selects the action included in that set; otherwise, it queries a human to disambiguate. These approaches ensure that the designed plans are correct with a user-defined probability. However, the larger this probability is, the larger the prediction sets are resulting in frequent human queries. The resulting help rate, i.e., the fraction of steps requiring user input, can become prohibitively high as LLMs are trained in an uncertainty-agnostic manner, with CP applied only post hoc \citep{huang2025rapid,chen2023longlora,zeng2025simplerl}.

We address this limitation by introducing  CoFineLLM (Conformal Finetuning for LLMs), the first uncertainty-aware fine-tuning framework for LLM-based planners that integrates CP during training. Our method aims to reduce prediction set sizes while preserving CP’s coverage guarantees. The key idea is to “simulate” conformalization during training by incorporating both calibration and prediction steps within mini-batches. We propose a novel loss function that combines a cross-entropy term with a CP-based term penalizing non-singleton prediction sets. To optimize this objective, we employ Low-Rank Adaptation (LoRA) \citep{hu2022lora} with a curriculum-based training scheme; however, other fine-tuning methods are compatible. During deployment, CP is applied to the fine-tuned model as in prior works to enable probabilistically correct decision-making \citep{ren2023robots,wang2024safe}. We show empirically that this approach significantly outperforms baselines in terms of help rates.

\textbf{Related Works:} \textit{(i) Conformal LLM-based Decision Making:} Prior works have applied CP to quantify uncertainty of LLMs used to address  planning, translation, and general MCQA problems \citep{ren2023robots,kumar2023conformal,su2024api,cherian2024large,vishwakarma2024prune,vishwakarma2024improving,liang2024introspective,campos2024conformal, lin2025domain,wang2024safe,wang2025conformalnl2ltl,wang2023conformal}. However, CP in these methods is used exclusively at test time, leading to frequent human interventions when prediction sets are non-singleton. A comprehensive survey on how CP has been used to quantify uncertainty of various AI-enabled components can be found in \citep{lindemann2024formal}. \textit{(ii): Conformal Training:} Several CP-aware training methods have been proposed for neural network classifiers \citep{pmlr-v128-colombo20a,bellotti2021optimized,stutz2021learning,noorani2024conformal,yan2024provably}. 
Efforts towards developing conformal finetuning methods for graph neural networks are presented in \citep{wang2025enhancing}. These works design differentiable loss functions to minimize prediction set size while preserving the CP coverage guarantees. However, unlike this paper, they focus on  models addressing single-step tasks (e.g., classification) and, therefore, target single-step prediction-set efficiency. Moreover, our empirical analysis suggests that existing CP-based loss functions, such as those proposed in \citep{stutz2021learning}, can be sensitive to the quality of the initial model and may not transfer effectively to LLMs. In particular, we observed empirically that when a pre-trained LLM (e.g., Gemma3-1B) is both incorrect and overconfident, these losses can produce near-zero gradients, effectively halting learning. As a result, such approaches may be less suitable for fine-tuning large models that begin far from convergence.
Building on these insights, we introduce a new CP-aware fine-tuning framework for sequential, multi-step decision-making tasks with LLMs. Our formulation explicitly mitigates the gradient sensitivity issue by defining a loss function that combines the standard cross-entropy objective with a novel CP regularizer. 

\textbf{Contribution:} The contribution of this paper can be summarized as follows: (i) We propose  CoFineLLM, the first CP-aware fine-tuning framework for LLMs in sequential decision-making tasks.
(ii) We instantiate this framework for robot planning and demonstrate substantial reductions in help rates from users compared to both uncertainty-agnostic  and conformal-aware fine-tuning methods.
(iii) We empirically show that the proposed framework is robust to the performance of the initial model, making it well suited for cases where the model is far from convergence.
(iv) We validate the method on hardware experiments, demonstrating robustness in out-of-distribution settings where test-time tasks and environments differ from those used during finetuning and calibration.
\vspace{-0.5cm}
\section{Problem Formulation}
\label{sec:problem}
\vspace{-0.2cm}

\textbf{Robot Agent and Environment.}
We consider a single robot equipped with a known skill set $\mathcal{S}=\{s_1,\ldots,s_S\}$ (e.g., `move', `turn', `pick up'). The robot operates in a known environment $\Omega \subset \mathbb{R}^d$, $d\in\{2,3\}$, containing $M>0$ semantic objects or regions $o_m$, $m\in\{1,\dots,M\}$. Each object $o_m$ is defined by its location and a semantic label (e.g., `box’, `key’). Applying a skill $s_j\in\mathcal{S}$ to an object $o_m$ yields an action $a(s_j,o_m)$; when it is clear from the context, it is simply denoted by $a$. The set of all valid actions $a$ is denoted by $\mathcal{A}$ and can be designed offline by enumerating admissible skill–object combinations.

\textbf{NL-based Planning Problem:}
The agent is assigned a mission $\phi$, expressed in NL, that can be completed if the robot executes a plan $\tau=\tau(1),\dots,\tau(t),\dots,\tau(H)$, defined as a finite sequence of actions $\tau(t)\in\mathcal{A}$, $t\in\{1,\dots,H\}$ for some horizon $H$. We formalize this setup by assuming a distribution $\mathcal{D}$ over mission scenarios $\xi_i = \{\phi_i, \mathcal{S}_i, H_i, \Omega_i\}$, where each scenario bundles together a mission $\phi_i$, the available skill set $\ccalS_i$, the mission horizon $H_i$, and the  environment $\Omega_i$. The distribution $\mathcal{D}$ is unknown, but we assume that i.i.d. scenarios can be drawn from it. When clear from context, we omit the subscript $i$.

\textbf{Conformal NL-based Planners:} Existing approaches can be employed to design plans $\tau$ that are correct with user-specified probability $1-\alpha\in(0,1)$ \citep{ren2023robots,wang2024safe}. These works frame the planning problem a sequence of MCQA problems solved by an LLM denoted by $g$. The MCQA problem at time step $t$ consists of a `question', denoted  by $\ell(t)$, providing a textual description of the scenario $\xi$ along with past actions (if any) the agent has taken, and the `choices' refer to the available actions $a\in\ccalA$ of the robot. The solution to this MCQA problem results in $\tau(t)$. To solve it, the confidence scores, denoted by $g(a|\ell(t))$, of the employed LLM for each  $a\in\ccalA$ are extracted. Using these scores, CP is applied to construct a prediction set defined as 
\begin{equation}\label{eq:predSet}
    \ccalC(t)=\{a \in \mathcal{A} \mid g(a | \ell(t)) \ge \delta \},
\end{equation}
where $\delta$ is the threshold computed offline (during a calibration phase; see Section \ref{sec:cp}) that depends on $1-\alpha$ and the model. If $\mathcal{C}(t)$ is a singleton, the corresponding action is executed; otherwise, the system queries a human to select the correct action if it lies in $\mathcal{C}(t)$, or halts otherwise. Repeating this procedure for all steps yields a plan $\tau$ that satisfies $\phi$ with probability at least $1-\alpha$. 

\textbf{Conformal Finetuning Problem:}
Our goal is to minimize the help rate, i.e., the fraction of time steps requiring human assistance, in conformal LLM-based planners through finetuning. Specifically, our goal is to develop an uncertainty-aware fine-tuning framework that explicitly reduces the size of the sets $\ccalC(t)$ generated at test time without compromising their coverage guarantees. 
During deployment, CP is applied to the finetuned model exactly as in standard conformal planners.

\vspace{-0.3cm}
\section{Method}
\label{sec:method}
\vspace{-0.2cm}

\textcolor{black}{In this section, we present  CoFineLLM, our conformal-aware finetuning framework for LLM-based planners. The key idea is to fine-tune the model using a novel loss function that penalizes prediction sets, generated via CP, that are non-singleton and fail to include the correct action. Constructing these sets (and consequently the loss) depends on the threshold $\delta$, introduced in \eqref{eq:predSet}, which varies as the model parameters are updated. Inspired by \cite{stutz2021learning}, we simulate the conformalization process during fine-tuning to dynamically adjust $\delta$. The subsections below provide a detailed overview of this process, summarized in Algorithm~\ref{alg:cpaware}. Section~\ref{sec:cp} first describes how prediction sets $\ccalC(t)$ in \eqref{eq:predSet} are constructed for a fixed model. Section~\ref{sec:loss} then presents our fine-tuning method, which optimizes the proposed loss function while periodically updating $\delta$, by invoking the procedure in Section~\ref{sec:cp}, to account for changes in the model parameters.}

\vspace{-0.2cm}
\subsection{Constructing Local Prediction Sets using CP}\label{sec:cp} 
\vspace{-0.1cm}

\begin{algorithm}[t]
\footnotesize
\caption{\texttt{ComputeConformalThreshold}}
\label{alg:calib}
\begin{algorithmic}[1]
\STATE \textbf{Input:} Calibration Dataset $\mathcal{M}_{\text{cal}}$; LLM $g$; Coverage level $1-\alpha$. \label{line:calib_input}


\STATE Compute nonconformity scores $\{\bar r_i\}_{i=1}^D$ on $\mathcal{M}_{\text{cal}}$ using LLM confidence scores $g$. \label{line:ncs}
\STATE Compute $\hat q = \text{Quantile}_{1-\alpha}(\{\bar r_i\})$ and define $\delta = 1 - \hat q$.
\label{line:quantile}
\STATE \textbf{Output:} Threshold $\delta$.
\end{algorithmic}
\end{algorithm}
\normalsize

In this section, we describe how the prediction sets $\ccalC(t)$ are constructed for a given mission scenario $\xi \sim \ccalD$, following the procedure in \citep{ren2023robots,wang2024safe}; see also Alg. \ref{alg:calib}. Constructing these sets requires two components:
(i) a calibration dataset comprising mission scenarios sampled from $\ccalD$ along with their corresponding correct robot plans, and
(ii) a nonconformity score (NCS) that quantifies the “error” of the LLM-based planner with respect to the ground truth.
We first outline the construction of the calibration dataset, followed by the definition of the NCS.

\textbf{Calibration dataset.}
We sample $D > 0$ i.i.d. scenarios $\xi_i \sim \mathcal{D}$. As discussed in Sec.~\ref{sec:problem}, generating a robot plan is framed as solving a sequence of MCQA problems. At each time step $t$, we construct the corresponding question (or prompt) associated with $\xi_i$, denoted by $\ell_{i,\text{cal}}(t)$, and identify the correct action $\tau_{i,\text{cal}}(t)$. This gives rise to the following sequence of prompts $\bar\ell_{i,\text{cal}} = \ell_{i,\text{cal}}(1), \ldots, \ell_{i,\text{cal}}(H_i)$  
\noindent and the corresponding sequence of correct actions (the ground-truth plan) $\tau_{i,\text{cal}} = \tau_{i,\text{cal}}(1), \ldots, \tau_{i,\text{cal}}(H_i).$
Repeating this process for all scenarios $\xi_i$ yields the calibration dataset: $\mathcal{M}_{\text{cal}} = \Big\{\big(\bar\ell_{i,\text{cal}}, \tau_{i,\text{cal}}\big)\Big\}_{i=1}^D$ .

\textbf{Non-conformity score (NCS).}
We define the NCS as the worst-case error of the LLM-based planner across all decision steps of a mission scenario. This process is summarized in Algorithm \ref{alg:calib}. Specifically, for a calibration scenario $\xi_i$, the NCS is given by
\begin{equation}\label{eq:NCS}
    \bar r_i = 1 - \min_{t \in \{1,\ldots,H_i\}} g\!\left(\tau_{i,\text{cal}}(t) \mid \ell_{i,\text{cal}}(t)\right),
\end{equation}
where $g(\tau_{i,\text{cal}}(t) \mid \ell_{i,\text{cal}}(t))$ denotes the confidence score assigned by the LLM $g$ to the correct action $\tau_{i,\text{cal}}(t)$ given the corresponding prompt $\ell_{i,\text{cal}}(t)$, as defined in Section~\ref{sec:problem}.  We compute the NCS $ \bar r_i$ for all calibration scenarios $\xi_i$ [line \ref{line:ncs}, Alg. \ref{alg:calib}]. The $1-\alpha$ empirical quantile $\hat q = \text{Quantile}_{1-\alpha}({\bar r_i})$ yields the threshold $\delta = 1 - \hat q$ used to form prediction sets [line \ref{line:quantile}, Alg.~\ref{alg:calib}].

\textbf{Constructing prediction sets.}
Consider an unseen test scenario $\xi_{\text{test}} \sim \mathcal{D}$ with planning horizon $H_{\text{test}}$ and unknown true plan. As discussed in Section~\ref{sec:problem}, planning is formulated as a sequence of MCQA problems. At each time step $t$, given the prompt $\ell_{\text{test}}(t)$ (associated with an MCQA problem of time step $t$), we define the prediction set $ \ccalC(t)$ introduced in \eqref{eq:predSet} which collects all actions whose confidence scores exceed the threshold $\delta$. This is done for all time steps $t \in \{1, \ldots, H_{\text{test}}\}$. As shown in \cite{ren2023robots,wang2024safe}, it holds that  $P(\tau_{\text{test}} \in \bar{\ccalC}) \ge 1-\alpha$, where $\tau_{\text{test}}$ is the true plan and $\bar{\mathcal{C}} = \mathcal{C}(1) \times \cdots \times \mathcal{C}(H_{\text{test}})$. A formal proof of this guarantee, along with a discussion of the case with multiple feasible solutions, can be found in \cite{ren2023robots}.

\vspace{-0.2cm}
\subsection{ CoFineLLM: Conformal Fine-Tuning for LLM-based Planners }
\label{sec:loss}
\vspace{-0.1cm}



In this section, we present our conformal-aware fine-tuning algorithm; see  Algorithm~\ref{alg:cpaware}. The algorithm takes as input: (i) an initial model $g$; (ii) a training dataset $\mathcal{M}_{\text{train}}$ (defined below); (iii) a calibration dataset $\mathcal{M}_{\text{cal}}$; (iv) a desired coverage level $1-\alpha$; and (v) hyperparameters $\lambda$ and $K$, introduced later. The calibration dataset $\mathcal{M}_{\text{cal}} = \{(\bar{\ell}_{i,\text{cal}}, \tau_{i,\text{cal}})\}_{i=1}^D$ consists of $D > 0$ pairs of prompt sequences and their corresponding ground-truth plans, constructed as described in Section~\ref{sec:cp} by sampling mission scenarios from $\mathcal{D}$. Note that this calibration dataset is not necessarily the same as the one used at test time. The training dataset $\mathcal{M}_{\text{train}} = \{(\bar{\ell}_{i,\text{train}}, \tau_{i,\text{train}})\}_{i=1}^T$ is generated in the same manner as the calibration dataset, using $T > 0$ mission scenarios sampled from $\mathcal{D}$.

The finetuning process runs until convergence where at each training epoch the model parameters, denoted by $\theta$, are updated to minimize a loss function $L(\theta,\delta)$ that will be introduced later. 
Given a threshold $\delta$, the model parameters $\theta$ are updated using stochastic gradient descent. Specifically, at each epoch, we sample from $\ccalM_{\text{train}}$ a mini-batch $\ccalB$ consisting of $B$ pairs $(\ell_{i,\text{train}}(t),\tau_{i,\text{train}}(t))$, i.e., prompts $\ell_{i,\text{train}}$ (i.e., questions in MCQA tasks) and their corresponding true actions $\tau_{i,\text{train}}(t) \in \ccalA$,  [line \ref{line:minibatch}, Alg. \ref{alg:cpaware}]. These individual prompts and actions can be drawn from different sequences $i$ in $\ccalM_{\text{train}}$ and do not need to correspond to the same time step $t$. This process repeats across epochs. As discussed earlier, the threshold $\delta$ used in the loss function $L(\theta,\delta)$ depends on the model parameters which evolve across epochs. To account for that, every $K$ epochs, Algorithm~\ref{alg:calib} is called to update $\delta$ using $\mathcal{M}_{\text{cal}}$ and the current parameters $\theta$ [line \ref{line:api2alg1}, Alg. \ref{alg:cpaware}].

Next, we define the loss function $L(\theta,\delta)$ that augments the standard cross-entropy objective with a conformal regularizer as follows:
\begin{equation}
\label{eq:finalobj-probtop1}
L(\theta,\delta)
= L_{\mathrm{CE}}(\theta)
+ \lambda\, L_{\mathrm{CP}}(\theta,\delta),
\qquad \lambda > 0,
\end{equation}
where $\lambda$ controls the trade-off between optimizing task accuracy and prediction set efficiency. The cross-entropy term $L_{\mathrm{CE}}= -\frac{1}{B} \sum_{j=1}^{B} \log g(y_j | \ell_j)$ 
increases the probability assigned to the ground-truth action $y$ across the datapoints $\ell$ included in $\ccalB$. 
The conformal term $L_{\mathrm{CP}}(\theta,\delta)$ is designed to minimize prediction set size only when the ground-truth action is already contained in the set. Intuitively, this encourages the model to first focus on improving prediction accuracy, when it is `wrong', and to optimize prediction set efficiency once it is `right'. This structure avoids conflicts between the two objectives, particularly during early training when the model’s predictions may be incorrect, and leads to a more stable finetuning process. We define the conformal loss as below: 

\begin{equation}
\label{eq:prob-top1-gated}
L_{\mathrm{CP}}(\theta,\delta)
= \frac{1}{|\mathcal{B}|} 
\sum_{(\ell, y)\in\mathcal{B}}
\psi(p_y, \delta)\,[p_{\max} - \delta]_+,
\end{equation}
where (i) $p_y = g(y \mid \ell)$ is the confidence score of the LLM assigned to the correct action $y$ associated with the prompt $\ell$; (ii) $p_{\max} = \max_{a \in \mathcal{A} \setminus \{y\}} g(a \mid \ell)$ is the largest probability assigned to any other action (called hereafter the `strongest rival action') [line \ref{line:probs}-\ref{line:cploss}, Alg. \ref{alg:cpaware}]; (iii) $[\cdot]_+$ is the ReLU function, i.e.,  $[a]_+= \max\{a,0\}$; and (iv) $\psi(p_y,\delta)=\mathbf{1}(p_y \ge \delta)$, i.e., $\psi(p_y,\delta)=1$ if $p_y \ge \delta$ and, therefore, the correct action is included in the prediction set, and $\psi(p_y,\delta)=0$ otherwise. By definition of $\psi(p_y,\delta)$, we have that $L_{\text{CP}}$ does not introduce penalties for those training examples $(\ell,y)\in\ccalB$ where $\psi(p_y,\delta)=0$, i.e., $y$ is not in the prediction set. Similarly, $L_{\text{CP}}$ introduces penalties only when (a) $\psi(p_y,\delta)=1$, i.e., $y$ is included in the prediction set and (b) $[p_{\max} - \delta]_+\geq 0$ meaning that the strongest rival action is included in the set (and, therefore, the set is non-singleton). In other words, as discussed earlier in this section, $L_{\text{CP}}$ introduces penalties only for non-singleton sets containing the true action. Then $\theta$ is update with the loss computed in \eqref{eq:finalobj-probtop1} [line \ref{line:param_update}, Alg. \ref{alg:cpaware}]. 

A challenge in the above definition of $L_{\text{CP}}$ is that it is not differentiable due to $\psi$. Thus, to enable gradient-based optimization, we use a smooth sigmoid approximation:
\[
\psi(p_y,\delta) = \sigma(T (p_y - \delta)),
\qquad \sigma(z)=\frac{1}{1+e^{-z}}.
\]
Observe that as $T \rightarrow \infty$, $\psi(p_y,\delta)$ converges to the above-mentioned step function $\mathbf{1}(p_y \ge \delta)$. 






\begin{algorithm}[t]
\footnotesize
\caption{CoFineLLM: Conformal Finetuning for LLM-based Planners 
}
\label{alg:cpaware}
\begin{algorithmic}[1]
\STATE \textbf{Input:} LLM $g$; Distribution $\mathcal{D}$; Coverage level $1-\alpha$; Weight $\lambda$; Period $K$. \label{line:input}

\STATE Sample scenarios from $\mathcal{D}$ and construct training dataset $\mathcal{M}_{\text{train}}$ and calibration dataset $\mathcal{M}_{\text{cal}}$ 
\label{line:obtain_dataset}

\REPEAT 
\STATE Sample minibatch $\mathcal{B}$ of step-level pairs $(\ell_{i,\text{train}}(t), \tau_{i,\text{train}}(t))$ drawn randomly from $\ccalM_{\text{train}}$.
\label{line:minibatch}
\STATE Every $K$ epochs, update $\delta \leftarrow \textsc{ComputeConformalThreshold}(g, \mathcal{M}_{\text{cal}}, 1-\alpha)$. \label{line:api2alg1}
\STATE Compute 
$p_{y} = g(y \mid  \ell)$ and 
$p_{\max} = \max_{a \in \mathcal{A} \setminus \{y\}} g(a \mid \ell)$ for all pairs $(\ell,y)\in\ccalB$. \label{line:probs}
\STATE Compute $L_{\mathrm{CP}}(\theta,\delta)
= \frac{1}{|\mathcal{B}|} 
\sum_{(\ell, y)\in\mathcal{B}}
\psi(p_y, \delta)\,[p_{\max} - \delta]_+$.
\label{line:cploss}
\STATE Update the model's parameters $\theta$ by minimizing $L(\theta,\delta)
= L_{\mathrm{CE}}(\theta)
+ \lambda\, L_{\mathrm{CP}}(\theta,\delta),
\qquad \lambda > 0$. \label{line:param_update}
    
\UNTIL{training converges} \label{line:end_repeat}

\STATE \textbf{Output:} Fine-tuned model $g$. \label{alg:line:output}
\end{algorithmic}
\end{algorithm}
\normalsize

\vspace{-0.4cm}
\section{Experiments}
\label{sec:experiments}
\vspace{-0.2cm}
In this section, we evaluate the effectiveness of CoFineLLM on language-instructed robot planning problems. Section \ref{subsec:env_tasks} describes the setup of our algorithm and baselines. Section \ref{subsec:main_results} presents comparative experiments showing that our method outperforms both uncertainty-agnostic and uncertainty-informed \citep{stutz2021learning} baselines across various metrics, such as user help rates. Section \ref{subsec:ablation} provides ablation studies demonstrating that CoFineLLM generalizes well to unseen test-time coverage thresholds differing from the one used during finetuning. Finally, Section \ref{subsec:hardware} empirically demonstrates the robustness of our method in out-of-distribution (OOD) hardware experiments.


\vspace{-0.2cm}
\subsection{Experiment Setup: Distribution of Mission Scenarios, Baselines, \& Evaluation Metrics}
\label{subsec:env_tasks}
\vspace{-0.1cm}

\begin{figure}[t]  
\centering
\includegraphics[width=0.55\linewidth]{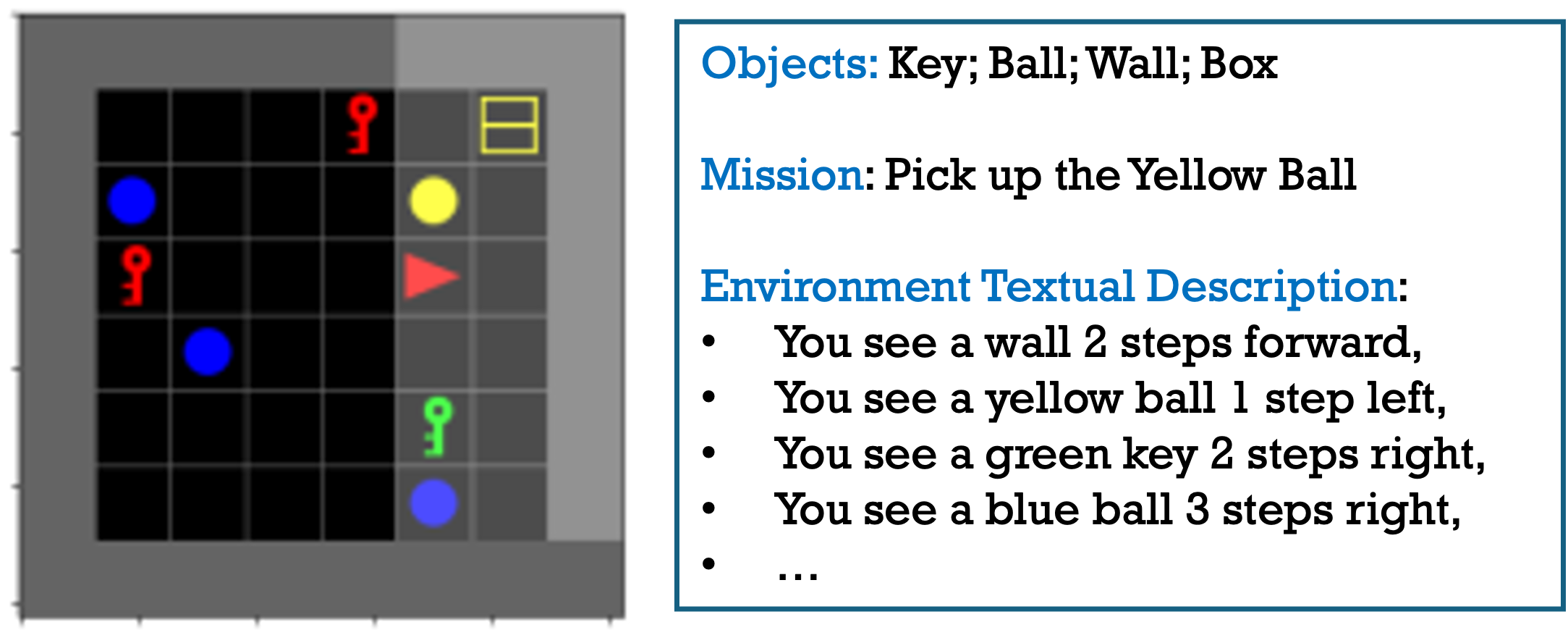}
\vspace{-5mm}
\caption{ 
Example environment from the BabyAI-Text simulator. The agent (red triangle) operates in a grid world with colored keys, balls, boxes, and walls, and receives NL mission (e.g., “pick up the yellow ball”). The simulator provides textual descriptions listing objects and their relative positions (e.g., “You see a yellow ball 1 step left”) for decision-making. 
}\vspace{-10mm}
\label{fig:babyai}
\end{figure}

\textbf{Distribution:} \textcolor{black}{We build on the BabyAI-Text environment~\citep{carta2023grounding}, a language-conditioned grid-world for grounded instruction following.} The distribution $\ccalD$ generates scenarios $\xi$ as follows. \textit{(a) Environment:} The distribution $\mathcal{D}$ produces random grid layouts using the BabyAI \citep{chevalier2018babyai} procedural generator. Each environment is an $8\times 8$ grid world composed of a $6\times 6$ navigable workspace surrounded by a one-cell-thick wall boundary (see Fig.~\ref{fig:babyai}). The outer wall cells are fixed and impassable, while the inner $6\times 6$ region contains the agent and objects. The environment is populated with objects belonging to four semantic categories: keys, balls, boxes, and doors. The number, color, and placement of objects for each category are sampled uniformly. Both the agent and the objects are initialized at random collision-free positions at the start of each episode. \textit{(b) Action Space}: The action space of the agent consists of six actions, i.e., $\mathcal{A} = \{a_i\}_{i=1}^6$, where $a_1 = \texttt{turn left}$, $a_2 = \texttt{turn right}$, $a_3 = \texttt{go forward}$, $a_4 = \texttt{pick up}$, $a_5 = \texttt{drop}$, $a_6 = \texttt{toggle}$. These six primitive actions enable the agent to move, manipulate objects, and interact with the environment. 
\textit{(c) Tasks}: Tasks are sampled uniformly from four categories of increasing complexity: (i) \textit{GoTo} — navigate to a randomly selected object in the environment;
(ii) \textit{PickUp} — navigate to and pick up a randomly selected object;
(iii) \textit{PickUpThenGoTo} — pick up one randomly selected object and then navigate to another randomly selected object;
(iv) \textit{PutNext} — pick up a randomly selected object and place it next to another randomly selected goal object. Each mission is expressed in natural language (e.g., “put the yellow ball next to the red key”). 

\noindent\textbf{Prompt Description:} As discussed in Section~\ref{sec:problem}, the prompt $\ell(t)$, corresponding to the question of the MCQA problem at time $t$, includes the following components:
(1) \textit{System description:} the symbolic action space and the decision objective;
(2) \textit{Environment description:} the relative position of the agent with respect to semantic objects at time $t$ (e.g., `You see a yellow ball 1 step left');
(3) \textit{Task description:} the natural-language mission $\phi$;
(4) \textit{Response format:} instructions specifying how the LLM should output its decision;
(5) \textit{Action history:} the sequence of past decisions up to time $t-1$; and
(6) \textit{Step index:} the current time step $t$.
Appendix \ref{app:prompt_example} shows an example prompt. 



\noindent\textbf{Training, Calibration, and Validation Datasets:} The training ($\ccalM_{\text{train}}$), calibration ($\ccalM_{\text{cal}}$), and validation datasets are constructed by sampling $4{,}000$, $400$, and $400$ mission scenarios, respectively, from  $\ccalD$. The training and calibration datasets are used during fine-tuning by Alg.~\ref{alg:cpaware}, while the validation dataset is reserved exclusively for evaluating the fine-tuned model.

\noindent\textbf{Models and Training Setup:}
In our evaluation, we consider the Gemma3-1B model~\citep{team2025gemma} finetuned using LoRA~\citep{hu2022lora} adapters while keeping the pretrained backbone frozen. This setup preserves the base model’s language understanding while enabling lightweight adaptation for language-instructed robot planning tasks.
To enhance the stability of fine-tuning, we also employ a \textit{phase-based curriculum learning} strategy that progressively introduces more complex mission categories while retaining a subset of simpler ones to mitigate catastrophic forgetting. This enables the LLM-based planner to first master short-horizon tasks before progressing to longer-horizon, compositional missions. All implementation details—including LoRA configuration, optimizer settings, learning rate, batch sizes, and phase schedules—are provided in Appendix~\ref{app:impl_details}. 
This fine-tuning framework is used to optimize our proposed loss function \eqref{eq:finalobj-probtop1}. 
Alternative fine-tuning frameworks could also be adopted.

\noindent\textbf{Evaluation Metrics.} We evaluate the finetuned model using four metrics:
\textit{(i) Set size:} the average size of prediction sets $\ccalC(t)$ across all validation scenarios.
\textit{(ii) Help rate:} the percentage of time steps $t$ across all validation scenarios where user help is requested (i.e., $|\ccalC(t)|>1$);
\textit{(iii) Coverage rate:} the percentage of validation scenarios in which the LLM-based planner generates a correct plan while asking for help from a user whenever $|\ccalC(t)|>1$ 
; and
\textit{(iv) Verification rate:} the percentage of validation scenarios for which the LLM-based planner designs a plan without asking for help  (i.e., all prediction sets at test time are singletons). All metrics are averaged over three independent runs.

\noindent\textbf{Baselines:} We consider two baselines to evaluate the effectiveness of our method. To ensure fair comparisons, both baselines share the same training setup and fine-tuning process described in Algorithm~\ref{alg:cpaware}; they differ from our approach only in the loss function they optimize. \textit{Uncertainty-agnostic (UA) baseline:}
This baseline optimizes only the cross-entropy loss, ignoring prediction set efficiency entirely. It can be recovered from our formulation in~\eqref{eq:finalobj-probtop1} by setting $\lambda = 0$.
\textit{Uncertainty-informed (UI) baseline:} 
This baseline optimizes the ConfTr loss introduced in \citep{stutz2021learning} for finetuning image classifiers. For each option (or label) $k$, a soft set-membership score is defined as $C_k(\theta) = \sigma((p_k - \delta)/\beta)$, where $p_k$ is the model’s predicted probability for option $k$ (that depends on the model's parameters $\theta$), $\delta$ is the confidence threshold generated by Alg. \ref{alg:calib}, $\beta$ controls the sharpness of the sigmoid $\sigma(\cdot)$, and $C_y(\theta)$ denotes the score for the ground-truth option $y$. Then, the ConfTr loss function is defined as $ L_{\mathrm{UI}}(\theta,\delta) = \log \big( (1 - C_y(\theta)) + \sum_{k \neq y} C_k(\theta) + \lambda \max(0, \sum_k C_k(\theta) - 1) \big) $ encouraging inclusion of the true class while penalizing unnecessarily large prediction sets. In our experiments, following~\citep{stutz2021learning}, we set $\beta = 0.1$.
Observe, that unlike our formulation, this loss does not include a cross-entropy term; instead, it simultaneously penalizes large prediction set sizes, even when the correct action is absent, and prediction sets that fail to contain the correct action. We also report the performance of the \textit{pre-trained} model.

\vspace{-0.3cm}
\subsection{Comparative Evaluation in In-Distribution Mission Scenarios}
\label{subsec:main_results}
\vspace{-0.1cm}
\begin{table}[t]
\setlength{\tabcolsep}{4pt}
\small
\centering
\caption{Comparative evaluations using $1{-}\alpha{=}0.95$ during both finetuning and validation. 
}
\label{tab:main_results}
\begin{tabular}{lcccc}
\toprule
\textbf{Method} 
& \textbf{Set Size}~$\downarrow$ 
& \textbf{Help Rate}~$\downarrow$ 
& \textbf{Coverage} 
& \textbf{Verification Rate}~$\uparrow$ \\
\midrule
Pre-trained 
& 6.000 & 100\% & 100\% & 0\% \\ 
UA 
& 1.138 & 12.7\% & 98.7\% & 41.2\%\\
Ours ($\lambda{=}0.1$)
& \textbf{1.109} (-2.54\%) 
& \textbf{9.7\%} (-23.67\%) 
& 98.7\% (+0\%) 
& \textbf{49.8\%} (+21.05\%) \\
\bottomrule
\end{tabular}\vspace{-4mm}
\end{table}


In this section, we present comparative numerical evaluations of our method and both baselines for a target coverage level of $1 - \alpha = 0.95$ during fine-tuning and validation; see Table~\ref{tab:main_results}. The loss functions of our method and the UI baseline are both configured using this coverage level. In our loss function (eq.~\ref{eq:finalobj-probtop1}), we set $\lambda = 0.1$. 
%
%
Before fine-tuning, the pre-trained model performs poorly on the considered planning tasks. For instance, its help rate is $100\%$, i.e., human assistance is required at every decision step across all validation missions to ensure a coverage rate of $95\%$; see the first row in Table~\ref{tab:main_results}. We also observed that the predictions of the pre-trained model are overconfident yet wrong in the sense that it assigned a large confidence score to a wrong action across all validation MCQA problems.  
After fine-tuning, CoFineLLM yields consistent improvements over both baselines across all evaluation metrics. Notably, relative to the UA baseline, the help rate decreases by $23.67\%$ while the verification rate increases by $21.05\%$, indicating a larger fraction of missions completed without human intervention; see Table~\ref{tab:main_results}.
\textcolor{black}{The UI baseline failed to learn any useful pattern and is stuck at 100\% help rate with an average prediction size of $6$ (i.e., $|\mathcal{A}|$). 
We observed empirically that this issue is not solely due to low initial accuracy but rather to the pretrained LLM producing overconfident yet incorrect predictions early in training. 
When a confidence score $p_k$ is either close to $1$ or $0$
, the corresponding $C_k$ quickly saturates near $1$ or $0$ under small changes in the model's parameters. 
Consequently, the loss landscape becomes nearly flat with respect to $\theta$, yielding vanishing gradients and preventing effective learning.}
%
%
%
This issue does not arise in our setup particularly because of the cross-entropy term in \eqref{eq:finalobj-probtop1}.
%
%
The coverage rate for all fine-tuned models remains close to the desired $1-\alpha=0.95$ across all baselines as expected due to the theoretical guarantees in \citep{ren2023robots,wang2024safe}.

\begin{table}[t]
\centering
\small
\setlength{\tabcolsep}{4pt}
\caption{Evaluation of finetuned model on unseen test-time coverage thresholds $1{-}\alpha$.
}
\begin{tabular}{c l cccc}
\toprule
$1{-}\alpha$ & Method & \textbf{Set Size$\downarrow$} & \textbf{Help Rate$\downarrow$} & \textbf{Coverage} & \textbf{Verification Rate$\uparrow$} \\
\midrule

\multirow{2}{*}{85\%}
& UA    & 1.037 & 3.7\% & 96.1\% & 76.3\% \\
& Ours  & \textbf{1.010} (-2.71\%) &
          \textbf{1.0\%} (-72.45\%) &
          \textbf{97\%} (+0.95\%) &
          \textbf{92.2\%} (+20.87\%) \\
\midrule

\multirow{2}{*}{90\%}
& UA    & 1.077 & 7.2\% & 97.3\% & 60\% \\
& Ours  & \textbf{1.046} (-2.76\%) &
          \textbf{4.4\%} (-39.20\%) &
          \textbf{97.6\%} (+0.26\%) &
          \textbf{70.3\%} (+17.22\%) \\
\midrule

\multirow{2}{*}{96\%}
& UA    & 1.159 & 14.38\% & 98.8\% & 37.8\% \\
& Ours  & \textbf{1.154} (-0.45\%) &
          \textbf{13.3\%} (-7.36\%) &
          \textbf{99\%} (+0.17\%) &
          \textbf{38.3\%} (+1.10\%) \\
\bottomrule
\end{tabular}\vspace{-4mm}
\label{tab:alpha_ablation}
\end{table}

\vspace{-0.2cm}
\subsection{Ablation Studies: Effect of Test-time Coverage Threshold $1-\alpha$ and Parameter $\lambda$ }
\label{subsec:ablation}
\vspace{-0.1cm}

\paragraph{Generalization to unseen test-time coverage level $1{-}\alpha$.} 
Here we evaluate the fine-tuned model from Section~\ref{subsec:main_results} (trained with $1{-}\alpha{=}0.95$ and $\lambda{=}0.1$) under test-time coverage targets $1{-}\alpha\in\{0.85,0.90,0.96\}$, which differ from the value used during fine-tuning. 
%
%
Comparative results against the finetuned model obtained in Section \ref{subsec:main_results} by the UA baseline are summarized in Table~\ref{tab:alpha_ablation}.
Our method consistently outperforms the UA baseline across all coverage levels. For example, at a lower coverage threshold ($1{-}\alpha{=}0.85$), the help rate decreases by $72.45\%$ and the verification rate improves by $20.87\%$. 
At a stricter coverage level of $1{-}\alpha{=}0.96$, it reduces help rate by $7.36\%$ and increases verification rate by $1.1\%$.
%
These results demonstrate that the benefits of conformal-aware fine-tuning persist even when the test-time coverage requirement differs from the one used during training, highlighting strong generalization of uncertainty calibration.


%

\begin{table}[t]
\centering
\small
\setlength{\tabcolsep}{4pt}
\caption{Evaluation under various values for $\lambda$ with $1{-}\alpha{=}0.95$ during finetuning and validation. 
}
\label{tab:lambda_ablation_expanded}
\begin{tabular}{lcccc}
\toprule
\textbf{Method} 
& \textbf{Set Size$\downarrow$} 
& \textbf{Help Rate$\downarrow$} 
& \textbf{Coverage} 
& \textbf{Verification Rate$\uparrow$} \\
\midrule

UA 
& 1.138 & 12.7\% & 98.7\% & 41.2\%\\
\midrule

$\lambda{=}0.5$
& \textbf{1.131} (-0.63\%) 
& \textbf{11.5\%} (-9.82\%) 
& \textbf{98.8\%} (+0.17\%) 
& \textbf{46.2\%} (+12.15\%) \\

$\lambda{=}0.3$
& \textbf{1.109} (-2.56\%) 
& \textbf{10.1\%} (-20.55\%) 
& 98.7\% (+0\%) 
& \textbf{48.7\%} (+18.27\%) \\

$\lambda{=}0.1$
& \textbf{1.109} (-2.54\%) 
& \textbf{9.72\%} (-23.67\%) 
& 98.7\% (+0\%) 
& \textbf{49.8\%} (+21.05\%) \\

\bottomrule
\end{tabular}\vspace{-4mm}
\end{table}

\paragraph{Sensitivity to $\lambda$.}
We fix the coverage level to $1{-}\alpha{=}0.95$ during both finetuning/calibration and validation, and vary the penalty coefficient $\lambda \in \{0.1, 0.3, 0.5\}$. As shown in Table~\ref{tab:lambda_ablation_expanded}, our method outperforms the UA baseline across all settings. A small penalty ($\lambda{=}0.1$) yields strong improvements, reducing help rate by $23.67\%$ and increasing verification rate by $21.05\%$. The best trade-off occurs at $\lambda{=}0.3$, which reduces prediction-set size by $2.56\%$, lowers help rate by $20.55\%$, and boosts verification rate by $18.27\%$. A larger penalty ($\lambda{=}0.5$) still improves verification but provides slightly weaker overall gains. Overall, these results show that our approach is robust to the choice of $\lambda$ and consistently improves prediction-set size and help rates.

\begin{figure}[t]
\centering
\includegraphics[width=1\linewidth]{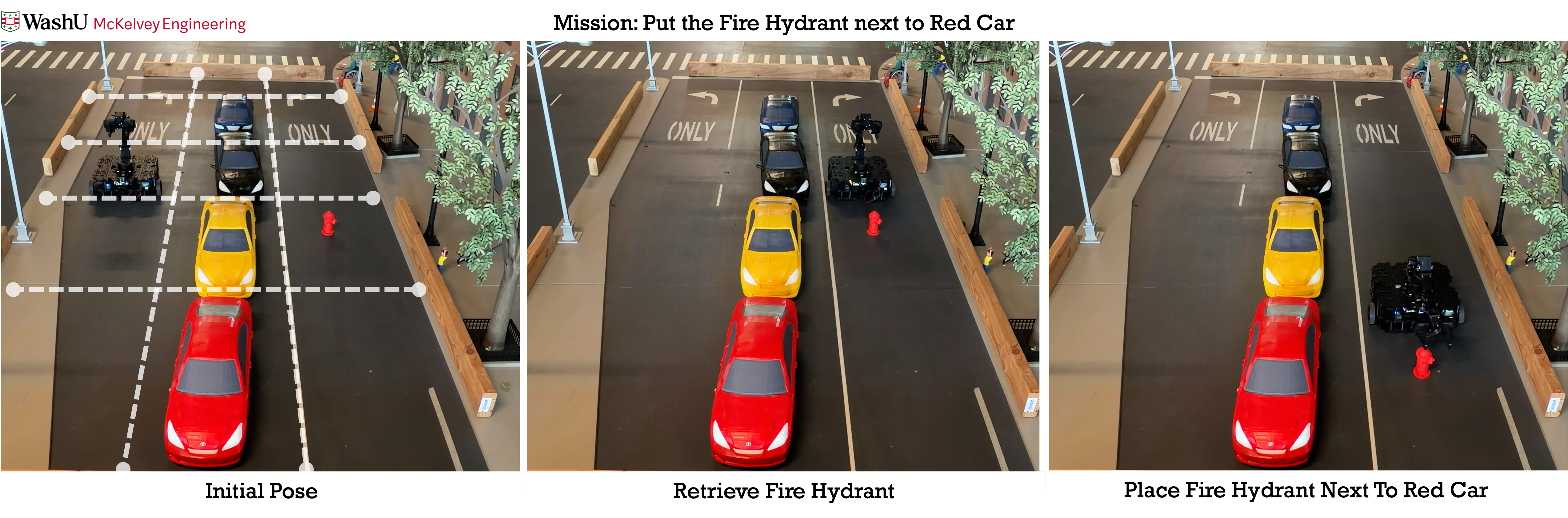}\vspace{-0.8cm}
\caption{ 
Hardware evaluation in a physical environment with out-of-distribution mission scenarios (sampled from $\ccalD'\neq\ccalD)$. 
The robot receives the NL instruction “put the fire hydrant next to the red car.” The left panel shows the $3{\times}5$ grid abstraction used for planning, where each cell represents a discrete navigation step. Using this representation, the LLM planner generates the action plan $\tau$, which directs the robot to turn, navigate to the hydrant, pick it up, move to the red car, and place the hydrant next to it.
}\label{fig:minicity}\vspace{-8mm}
\end{figure}

\begin{table}[t]
\setlength{\tabcolsep}{4pt}
\small
\centering
\caption{Evaluation in OOD scenarios using $1{-}\alpha{=}0.95$ during finetuning and validation.
}
\label{tab:hardware_results}
\begin{tabular}{lccccc}
\toprule
\textbf{Method}  
& \textbf{Set Size}~$\downarrow$ 
& \textbf{Help Rate}~$\downarrow$ 
& \textbf{Coverage} 
& \textbf{Verification Rate}~$\uparrow$ \\
\midrule
UA 
& 1.131 & 11.7\% & 95.1\% & 58.9\% \\
Ours ($\lambda{=}0.1$)
& \textbf{1.072} (-5.22\%) 
& \textbf{6\%} (-48.67\%) 
& \textbf{98.1\%} (+3.16\%) 
& \textbf{71.7\%} (+21.70\%) \\
\bottomrule
\end{tabular}\vspace{-5mm}
\end{table}


\vspace{-0.3cm}
\subsection{Hardware Experiments \& Comparative Evaluations in OOD  Mission Scenarios}
\label{subsec:hardware}
\vspace{-0.1cm}

\noindent\textbf{OOD Validation Scenarios:} In this section, we evaluate the finetuned model obtained in Section \ref{subsec:main_results}  on OOD mission scenarios. Specifically, we consider cases where validation scenarios are sampled from a distribution $\ccalD'$ that differs from the distribution $\ccalD$, used during calibration and finetuning, in the environments it generates. The distribution $\ccalD'$ generates $3{\times}5$ grid-world environments populated with traffic cones, fire hydrants, and cars, where cars may appear in one of six possible colors. 
Both $\ccalD$ and $\ccalD'$ generate tasks from the same four mission categories, but the tasks are defined over different objects, as the environments generated by these distributions differ. The distribution $\ccalD'$ is not accessible and thus cannot be used for calibration or finetuning. Thus, calibration data during both finetuning and test time, the LLM-based planner are sampled from $\ccalD$ while validation mission scenarios are generated by $\ccalD'$. The coverage threshold during both finetuning and test time is set to $0.95$.

\noindent\textbf{Comparative Results:} The comparative results over 80 validation mission scenarios generated by $\ccalD'$ against the UA baseline are summarized in Table \ref{tab:hardware_results}. 
%
Using the threshold $\delta$ calibrated using data from $\mathcal{D}$, CoFineLLM substantially improves help and verification rates. 
Specifically, the verification rate  rises by +21.7\%  and help rate drops by 48.67\% compared to the UA baseline. Interestingly, models finetuned by both methods preserve the $1-\alpha$ coverage level despite the distribution shift. 

\noindent\textbf{Hardware Demonstration:} The plans generated by the finetuned LLM-based planner are demonstrated on real robot platform TurtleBot3 Waffle-Pi equipped with an OpenManipulator-X arm~\citep{emanual_turtlebot} in a miniature urban environment modeled as $3\times 5$ grid world, populated with traffic cones, fire hydrants, and cars. Snapshots from such a demonstration are provided in Fig. \ref{fig:minicity}.

\vspace{-0.7cm}
\section{Conclusion}
\vspace{-0.2cm}
This paper introduced the first conformal-aware finetuning framework for LLMs used for language-instructed robot planning. Our method reduces the size of prediction sets generated by CP at test time, thereby decreasing the need for user intervention. Our experiments demonstrate that the proposed method consistently outperforms baselines in terms of prediction-set size and help rate.

\acks{This work was supported in part by the NSF award CNS $\#2231257$.}

\bibliography{YK_bib}

\appendix
\begin{appendices}

\section{Example Prompt}
\label{app:prompt_example}

For completeness, we show an example of the full prompt used to query the LLM planner during fine-tuning and evaluation. 
Each prompt encodes (i) the action space and response format, (ii) the mission goal, (iii) the current textual observation, 
and (iv) the action history up to the current step. 
This structure transforms each decision point into a multiple-choice question where the LLM selects one symbolic action.

\begin{quote}
\small
\texttt{Select an action by its corresponding number: \\
0: turn left \\
1: turn right \\
2: go forward \\
3: pick up object \\
4: drop object \\
5: toggle \\
\\
Goal of the agent: go to a red ball after you pick up the grey key \\
\\
Observation: You see a wall 2 steps back and 1 step left. You see a wall 1 step back and 1 step left. You see a wall 1 step left. You see a wall 1 step forward and 1 step left. You see a wall 2 steps forward and 1 step left. You see a wall 3 steps forward and 1 step left. You see a wall 4 steps forward and 1 step left. You see a wall 5 steps forward and 1 step left. You see a wall 2 steps back. You see a wall 5 steps forward. You see a wall 2 steps back and 1 step right. You see a red ball 4 steps forward and 1 step right. You see a wall 5 steps forward and 1 step right. You see a wall 2 steps back and 2 steps right. You see a yellow key 1 step forward and 2 steps right. You see a yellow ball 2 steps forward and 2 steps right. You see a grey key 3 steps forward and 2 steps right. You see a wall 5 steps forward and 2 steps right. You see a wall 2 steps back and 3 steps right. You see a red ball 1 step forward and 3 steps right. You see a green key 4 steps forward and 3 steps right. You see a wall 5 steps forward and 3 steps right. You see a wall 2 steps back and 4 steps right. You see a red key 1 step back and 4 steps right. You see a wall 5 steps forward and 4 steps right. You see a wall 2 steps back and 5 steps right. You see a blue box 5 steps right. You see a green box 2 steps forward and 5 steps right. You see a red ball 3 steps forward and 5 steps right. You see a wall 5 steps forward and 5 steps right. You see a wall 2 steps back and 6 steps right. You see a wall 1 step back and 6 steps right. You see a wall 6 steps right. You see a wall 1 step forward and 6 steps right. You see a wall 2 steps forward and 6 steps right. You see a wall 3 steps forward and 6 steps right. You see a wall 4 steps forward and 6 steps right. You see a wall 5 steps forward and 6 steps right. \\
\\
Previous Steps: \\
\\
Your next action (choose number): \\
Action:
}
\end{quote}

This example corresponds to a mid-episode decision in the \textit{PickUpThenGoTo} mission family. 
During training, only the observation and history fields evolve over time, 
while the system description and response format remain constant across steps.

\section{Implementation Hyperparameters}
\label{app:impl_details}

\begin{table}[h]
\centering
\footnotesize
\caption{Key hyperparameters used in CP-aware fine-tuning.}
\label{tab:hyperparams}
\begin{tabular}{ll}
\toprule
\textbf{Hyperparameter} & \textbf{Value} \\
\midrule
Temperature ($T$) & Hard Indicator $\mathbf{1}(p_y \ge \delta) $\\
Period between updates of $\delta$ ($K$) & 10 \\
Batch Size & 4\\
Learning rate & 5e--5 \\
Optimizer & Adam \\
LoRA rank ($r$) & 16 \\
LoRA scaling  & 128 \\
Coverage level ($1-\alpha$) & 0.95 \\
Phase start epochs & [1, 6, 11, 21] \\ Retained samples per phase & [100, 100, 500, 1000] \\
\bottomrule
\end{tabular}
\end{table}

Table~\ref{tab:hyperparams} summarizes the core hyperparameters used for CP-aware fine-tuning. We train with Adam at a learning rate of $5\times10^{-5}$ and a batch size of $4$, using LoRA adapters (rank $16$, scaling $128$). We target $95\%$ conformal coverage and introduce the CP loss in phases, gradually increasing the number of calibration samples retained during training. 

\end{appendices}

\end{document}